\pdfoutput=1
\documentclass[11pt]{article}

\usepackage[]{acl}
\usepackage{graphicx}

\usepackage{times}
\usepackage{latexsym}

\usepackage[T1]{fontenc}
\usepackage[utf8]{inputenc}

\usepackage{microtype}

\title{UPB at SemEval-2022 Task 5: Enhancing UNITER with Image Sentiment and Graph Convolutional Networks for Multimedia Automatic Misogyny Identification}

\author{Andrei Paraschiv, Mihai Dascalu, Dumitru-Clementin Cercel\\
   University Politehnica of Bucharest, Faculty of Automatic Control and Computers\\
   \{andrei.paraschiv74, mihai.dascalu, dumitru.cercel\}@upb.ro}

\begin{document}
\maketitle
\begin{abstract}
In recent times, the detection of hate-speech, offensive, or abusive language in online media has become an important topic in NLP research due to the exponential growth of social media and the propagation of such messages, as well as their impact. Misogyny detection, even though it plays an important part in hate-speech detection, has not received the same attention. In this paper, we describe our classification systems submitted to the SemEval-2022 Task 5: MAMI - Multimedia Automatic Misogyny Identification. The shared task aimed to identify misogynous content in a multi-modal setting by analysing meme images together with their textual captions. To this end, we propose two models based on the pre-trained UNITER model, one enhanced with an image sentiment classifier, whereas the second leverages a Vocabulary Graph Convolutional Network (VGCN). Additionally, we explore an ensemble using the aforementioned models. Our best model reaches an F1-score of 71.4\% in Sub-task A and  67.3\% for Sub-task B positioning our team in the upper third of the leaderboard. We release the code and experiments for our models on GitHub\footnote{https://github.com/readerbench/semeval-2022-task-5}.

\end{abstract}

\section{Introduction}

The web and social network platforms, in particular, have become a significant part of our modern social lives. Sharing information, opinions, news, and jokes through these platforms with friends and family are now daily routines. One of the most prevalent forms of jokes in  social networks are memes. Internet memes are small cultural units that are transformed, mixed, and shared using online platforms, often spreading in a viral manner \cite{Milner}. A prolific part are image-based memes, often available as templates that are accompanied by humorous or witty text. Unfortunately, a considerable proportion of the memes shared by Internet users are offensive or even hateful messages\footnote{https://www.hmc.org.uk/blog/third-teenage-boys-admit-sending-receiving-racist-homophobic-content-online/}.  

Detecting hate and offensive speech is a significant task for any online platform. Not only companies have this legal obligation in most countries, but also such language establishes a toxic environment that is detrimental to any online community on the long run. Hate speech can take multiple forms, but it is most frequently encountered as a disparaging message on the basis of a characteristic as race, gender, religion, and other criteria; Misogyny is one such frequent form specific to meme culture \cite{Drakett_Rickett_Day_Milnes_2018, Phillips_2012}. Detecting hate speech is often a hard problem, even in an uni-modal setting since the message often relies on the context, addresses current events, and incorporates cultural knowledge that cannot be easily incorporated into an automated model. The multi-modality of Internet memes increases the difficulty of the task since many memes can have a seemingly benign text that, contextualized with the associated image, becomes offensive or hateful. 

Multi-modal hate speech detection had less attention in the research literature than traditional text-only methods. In the past two years, several datasets and challenges have addressed this by proposing detection tasks on meme-based data \cite{kiela2020hateful, Gasparini_Rizzi_Saibene_Fersini_2021, Miliani_Giorgi_Rama_Anselmi_Lebani_2020}. Misogyny detection, as a subgroup of hate speech detection tasks, has also been more frequently encountered in research in recent years. The series of Automatic Misogyny Identification tasks proposed at IberEval 2018, EVALITA 2018, EVALITA 2020, and TRAC-2020 \cite{fersini2018overviewEvalita,fersini2018overview,fersini2020ami, trac2-report, trac2-dataset} focused on the classification of tweets in English, Spanish, Italian, Bangla, and Hindi languages. In these tasks, researchers identified misogynous tweets and classified them as aggressive/non-aggressive or active/passive. 

Semeval-2022 Task 5: MAMI - Multimedia Automatic Misogyny Identification \cite{mami2022semeval} is a multi-modal classification task, aiming to detect misogynous memes by leveraging both image and text information. The task includes two sub-tasks: a binary identification of misogynous/non-misogynous memes (Sub-task A), and a multi-label classification distinguishing between the types of misogyny, namely: stereotype, shaming, objectification, and violence (Sub-task B).

In this paper, we present our contribution to this task by proposing two architectures based on the pre-trained multimodal UNITER (UNiversal Image-TExt Representation) model \cite{Chen_Li_Yu_Kholy_Ahmed_Gan_Cheng_Liu_2020}, as well as an ensemble from these two models. UNITER is an early fusion model pre-trained on large text-image datasets. UNITER leverages visual and location features extracted with Faster R-CNN \cite{Ren_He_Girshick_Sun_2016}, together with WordPiece encodings \cite{wu2016google} derived from text tokens using a Transformer-based model \cite{vaswani2017attention}. UNITER learns a generalizing representation for the text-image context by bringing the visual and text representations in a common embedding space. We use UNITER at the core of our two architectures: the first one enhances the pre-trained model with image sentiment features using a VGG-19 model \cite{Vadicamo_Carrara_2017}, while the second leverages graph convolutions on a co-occurrence graph \cite{kipf2016semi} built from an external dataset. 

\section{Background}

Multi-modal tasks were traditionally associated with visual question answering \cite{goyal2017making}, image captioning \cite{gurari2020captioning}, audio-visual speech recognition \cite{paraskevopoulos2020multiresolution}, or cross-modal retrieval \cite{wang2016learning}. With success of competitions like the Hateful Memes Challenge \cite{kiela2020hateful}, more research focused on multi-modal offensive classification. Pre-trained transformer models such as ViLBERT \cite{lu2019vilbert}, VisualBERT \cite{li2019visualbert}, LXMERT \cite{tan2019lxmert}, Oscar \cite{Li_Yin_oscar_2020}, and others, dominated the competition leaderboard, either as stand-alone models or in large ensembles. While considering UNITER, \newcite{Lippe_Holla__2020} used an ensemble that placed them in the top 5 teams. 

\section{Method}
We explored two ways of enhancing UNITER, first by adding a unimodal late fusion with a visual sentiment classifier, and second by using a multimodal early fusion with a modified Vocabulary Graph Convolutional Network (VGCN) \cite{Lu_Du_Nie_2020, paraschiv2021graph}.

\subsection*{Dataset analysis and preprocessing}
The training dataset contains 10,000 records, half of them misogynous. One misogynous record can have one or more of the four labels. The class distribution among the types of misogyny is as follows:  shaming - 1,274, stereotype - 2,810, objectification - 2,208 and violence - 953 samples. 

The text modality from the dataset was obtained, as far as we can tell, through OCR without any manual cleaning. This lead to the inclusion of date, times, mobile carrier names, Facebook user names or words on some unrelated objects in the image like \textit{"Verizon LTE 4:41 PM Bikram Dec 11 at 12:31 AM"}. Additionally, many memes contain the watermark of the publishing website: "imgflip.com", "makeameme.org", "memez.com". Since most memes use full uppercase fonts, the letter casing was not reliable throughout the dataset and we choose to lowercase all training entries. 

In our experiments, we tried two types of data cleaning techniques: one where we remove all timestamps and date mentions using the SUTime library \cite{chang-manning-2012-sutime} and a second  with the supplementary step of removing all website mentions and Twitter usernames from the text. 
\begin{figure*}[htp!]
    \centering
    \includegraphics[width=13cm]{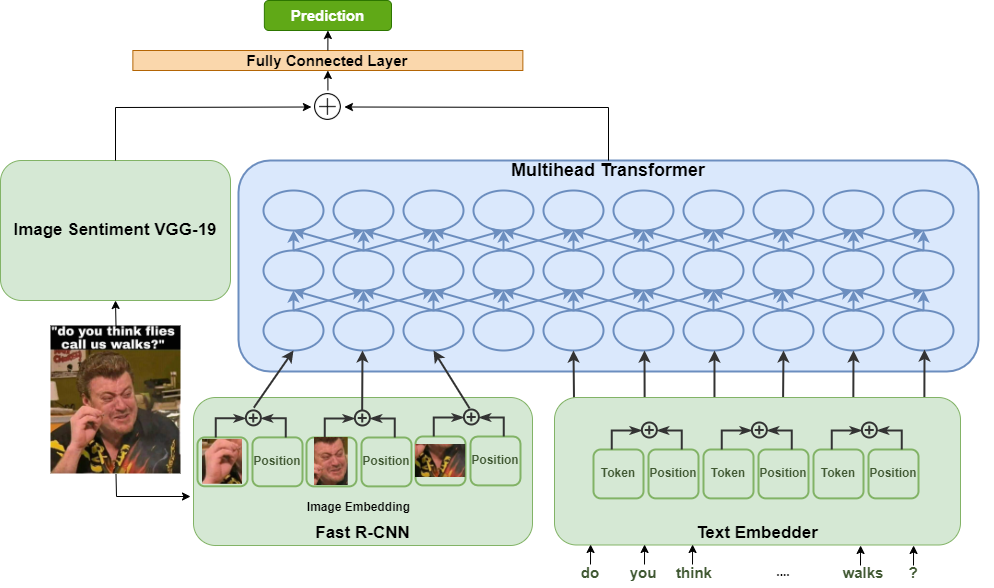}
    \caption{Illustration of the proposed UNITER-Sentiment model.}
    \label{fig:model-sentiment}
\end{figure*}

\subsection*{Visual Sentiment-enhanced UNITER}

Offensive texts, hate speech, and misogynous language are often correlated with negative sentiments, whereas the tone, context, and content is often highly loaded with polarized language \cite{Ali_Ehsan-Ul-Haq_Rauf_Javed_Hussain_2021, Gitari_Zhang_Damien_Long_2015}; as such, our intuition to enhance the UNITER model with a sentiment classifier. We focused only on  image modality since large language models often already capture features required for text sentiment analysis. All images were classified using a pre-trained VGG-19 model \cite{simonyan2014very} fine-tuned on the T4SA dataset \cite{Vadicamo_Carrara_2017}. The resulting 4,096 sized feature vectors were fused by concatenation with the UNITER's pooled output and classified through a fully connected layer in the classes for each sub-task (see Figure \ref{fig:model-sentiment}).  

\subsection*{VGCN-enhanced UNITER}

Graph enhanced BERT models have proven to be powerful for text classification \cite{mamani2021aggressive}, even in multimodal tasks \cite{vlad2020upb}. Using a GCN on a vocabulary graph \cite{kipf2016semi}, the model can train an embedding layer to be fused with the Transformer embedding, before being processed by the multi-head attention layers. For our architecture, we employ a novel approach,  namely to create a heterogeneous graph with nodes that represent text tokens and objects detected by the R-CNN layer, in the corresponding image.

\begin{figure*}[htp]
    \centering
    \includegraphics[width=13cm]{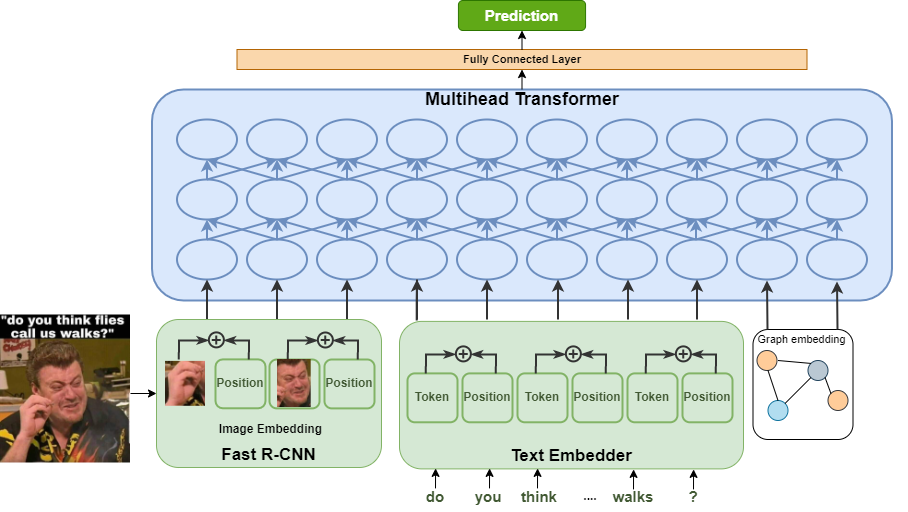}
    \caption{Illustration  of the proposed UNITER-VGCN model.}
    \label{fig:model-vgcn}
\end{figure*}

A Kaggle dataset\footnote{https://www.kaggle.com/zacchaeus/meme-project-raw} containing 3,000 meme templates and their various possible text captions, totaling 533,827 text records, was used to build the aforementioned graph. The same R-CNN layer and object-token encoding as the UNITER model were considered to create a co-occurrence graph, having nodes as BERT-token-IDs and detected object-IDs,  while edge values were computed using Point-wise Mutual Information (PMI). In contrast to \cite{Lu_Du_Nie_2020}, the obtained graph is independent of the training dataset and can be used for several tasks in the same domain.

In the training step, the UNITER image and text embeddings are fed through a GCN layer on top of the pre-built graph, thus generating a new embedding vector. The concatenated text, image, and graph embeddings are then processed by multi-head attention layers, while the pooled outputs are classified using a fully connected layer (see Figure \ref{fig:model-vgcn}).

\subsection*{Ensemble model}

In order to leverage the learning of both proposed models, we also utilize an ensemble formed from multiple versions of both models, trained on different train/dev/validation splits, that are then combined via a soft voting scheme. The best performing trained versions evaluated on our test set were picked in the ensemble.  For our final submission, we used the votes from 2 UNITER-Sentiment with UNITER-base, 2 with UNITER-large, and 7 UNITER-VGCN with UNITER-large. We chose these components of the ensemble based on the results for Sub-task B on our validation set, as seen in Table \ref{tab:resultsValidation}. Details on general hyper-parameters across all models are presented in the subsequent section. For UNITER-Sentiment models, we used 150 warm-up steps with a weight decay of 0.01, in contrast to only 120 warmup steps with a weight decay of 0.1 for UNITER-VGCN models. All UNITER-VGCN models in our ensemble had a graph embedding size of 16 and the edges had a minimum normalized PMI \cite{bouma2009normalized} of 0.3. Since we trained each of the 11 ensemble components on a different train/dev/validation partition with the same seed, each configuration converged on different weights, with different performances. The results of all 11 models were combined through a soft-voting scheme based on the average predicted probability for each label. Additional configurations were trained, but overall this ensemble had the best test performance among our submissions to SemEval-2022 Task 5.

\begin{table*}[ht!]
\centering
\begin{tabular}{lp{12mm}p{11mm}p{21mm}|p{12mm}p{11mm}p{21mm}}
\hline
 &  & Sub-task A &  &  &Sub-task B & \\
\hline
\textbf{Model} & \textbf{Precision} & \textbf{Recall} & \textbf{Weighted-F1} & \textbf{Precision} & \textbf{Recall} & \textbf{Weighted-F1}  \\
\hline
UNITER-base+Sentiment$_1$ & 81.60\% & 63.95\%  & 67.17\% & 68.41\% & 42.13\% & 61.34\%\\
UNITER-base+Sentiment$_2$ & 83.80\% & 62.82\% & 66.16\% & 63.59\% & 44.72\% & 63.16\%\\
UNITER-large+Sentiment$_1$ & 83.00\% & 63.17\% & 66.47\% & 57.67\% & 49.68\% & 64.69\%\\
UNITER-large+Sentiment$_2$ & \textbf{86.80}\% & 62.54\% & 66.13\% & 70.51\% & 45.00\% & 64.18\%\\
UNITER-large+VGCN$_1$   &  78.80\% & 65.78\% & 68.59\% & 63.39\% & 48.87\% & \textbf{65.89}\%\\
UNITER-large+VGCN$_2$   &  78.40\% & 64.05\% & 66.78\% & 58.38\% & 47.06\% & 64.50\%\\
UNITER-large+VGCN$_3$   &  78.40\% & \textbf{68.89}\% & \textbf{71.36\%} & 66.30\% & 46.54\% & 64.84\%\\
UNITER-large+VGCN$_4$   &  77.60\% & 65.10\% & 67.70\% & 48.24\% & \textbf{51.84}\% & 64.25\%\\
UNITER-large+VGCN$_5$   &  86.60\% & 63.03\% & 66.74\% & \textbf{70.91}\% & 42.76\% & 61.66\%\\
UNITER-large+VGCN$_6$   &  79.20\% & 68.39\% & 71.12\% & 58.48\% & 50.62\% & 65.35\%\\
UNITER-large+VGCN$_7$   &  81.60\% & 65.18\% & 68.50\% & 53.66\% & 50.18\% & 65.56\%\\
\hline
\end{tabular}
\caption{Results on our validation set for the Ensemble components.}

\label{tab:resultsValidation}
\end{table*}
\subsection{Experimental Setup}

\begin{table*}[ht!]
\centering
\begin{tabular}{lp{21mm}p{21mm}p{27mm}}
\hline
\textbf{Model} & \textbf{Precision} & \textbf{Recall} & \textbf{Weighted-F1}  \\
\hline
UNITER-base+Sentiment$_1$ \hspace{2mm}  & 72.60\% & 67.60\% & 68.86\% \\
UNITER-base+VGCN \hspace{2mm}  & \textbf{86.20}\% & 61.66\% & 64.91\% \\
\hline
UNITER-large+Sentiment$_1$ \hspace{2mm}  & 83.00\% & 63.16\% & 66.47\% \\
UNITER-large+VGCN$_3$   &  78.40\% &  \textbf{68.89}\% & \textbf{71.36}\%  \\
\hline
Ensemble           & 83.20\% & 66.88\%  & 70.56\%  \\
\hline
\end{tabular}
\caption{Results on the official test set for Sub-task A.}

\label{tab:resultsA}
\end{table*}

\begin{table*}[ht!]
\centering
\begin{tabular}{lp{21mm}p{21mm}p{27mm}}
\hline
\textbf{Model} & \textbf{Precision} & \textbf{Recall} & \textbf{Weighted-F1}  \\
\hline
UNITER-base+Sentiment$_2$ \hspace{2mm}  & 59.97\% & 46.43\% & 63.99\% \\
UNITER-base+VGCN \hspace{2mm}  & 63.99\% & 45.61\% & 63.39\% \\
\hline
UNITER-large+Sentiment$_1$ \hspace{2mm}  & 57.67\% & 49.68\% & 64.68\% \\
UNITER-large+VGCN$_1$   &  \textbf{66.30}\% &  46.54\% & 64.84\%  \\
\hline
Ensemble           & 63.19\% & \textbf{50.65}\%  & \textbf{67.31\%}  \\
\hline
\end{tabular}
\caption{Results on the official test set for Sub-task B.}

\label{tab:resultsB}
\end{table*}

\begin{table*}[ht!]
\centering
\begin{tabular}{l|p{19mm}c|p{19mm}c}
\hline
 \textbf{Rank} & \textbf{Team} & \textbf{Sub-task A Score}  & \textbf{Team} & \textbf{Sub-task B Score}  \\
\hline
1 & SRC-B & 83.4\% & SRC-B & 73.1\% \\
2 & DD-TIG & 79.4\% & TIB-VA & 73.1\% \\
3 & beantown & 77.8\% & PAFC & 73.1\% \\
\hline
 & UPB (our) & 71.4\% & UPB (our) & 67.3\% \\
\hline
 & Baseline & 65.0\% & Baseline & 62.1\% \\ 
\hline
\end{tabular}
\caption{Comparison for Sub-tasks A and B between the top 3 team results, our scores, and the competition baseline.}

\label{tab:resultsTeams}
\end{table*}

Since the regions of interest (ROI) detection in the R-CNN layer\footnote{https://github.com/MILVLG/bottom-up-attention.pytorch} returns the probability for each rectangle, in our experiments, we use the minimum threshold for a ROI to be included in the dataset as an input hyperparameter. We experimented with both versions of UNITER: \textit{UNITER-base} with 12 attention heads and 768 output dimensions, and \textit{UNITER-large} with 24 attention heads and 1024 dimensions\footnote{https://github.com/ChenRocks/UNITER}. In order to mitigate class imbalance from Sub-task B, we used a weighted binary cross entropy loss using the class distribution in the training set. Other hyperparameter values were determined through grid search. Thus, the minimum confidence level for ROIs was set at 0.7, the learning rate at 1e\textsuperscript{-4} using an AdamW optimizer \cite{loshchilov2017decoupled}, and the maximum text length at 64 tokens. We experiment with GCN embedding sizes between 8 and 32, and selected 16 as the optimum value. 

In order to further address the class imbalance between the misogyny sub-types, we experimented with oversampling from the minority classes, as an alternative to the loss re-weighting technique. Also, from our experiments, we learned that using an additional "non-misogynous" class besides the four misogyny types improved the model performance. 

During training, we optimize the weighted-average F1-score (i.e., the F1-scores computed for each label, afterwards weighted by the label support) using early-stopping with a patience of 2 epochs. 

\section{Results}
While considering the two proposed data preprocessing techniques, removing the websites from the textual information proved to decrease the performance of the models. Even though removing the source from the texts would provide a better generalization in a production setting, this information turned out to be a clue on misogyny content in this dataset. 

Compensating for the unbalanced classes with oversampling from the minority classes proved to be less impactful than weighting the loss of the positive classes. Also, we noticed a strong tendency to overfit during our experiments. A common used mitigation technique is to augment the training data with small variations - e.g., data augmentation using similar words in the embedding space \cite{wang-yang-2015-thats}, as well as simpler replacement, swap, and deletion methods \cite{wei-zou-2019-eda}. None of the applied augmentation methods improved performance. We thus hypothesize that although these augmentations create similar meanings, innuendos get lost. For example, the phrase \textit{"Her: Excuse me, I'm trying to put a \textbf{load} in the dishwasher Him: Same @gogomeme"} would get an augmented counterpart \textit{"Her: Excuse me, I'm trying to put a \textbf{\textit{burdened}} in the dishwasher Him: Same @gogomeme"}. Changing the word "load", which in the context has a double meaning, loses the implicit misogynistic comparison of the woman to a dishwasher. Similarly, a text like \textit{"my horny girlfriend on her period me"} augmented as \textit{"my horny girlfriend on her deadline me"} would not improve the training.

\begin{figure*}[htp]
    \centering
    \includegraphics[width=3.1cm]{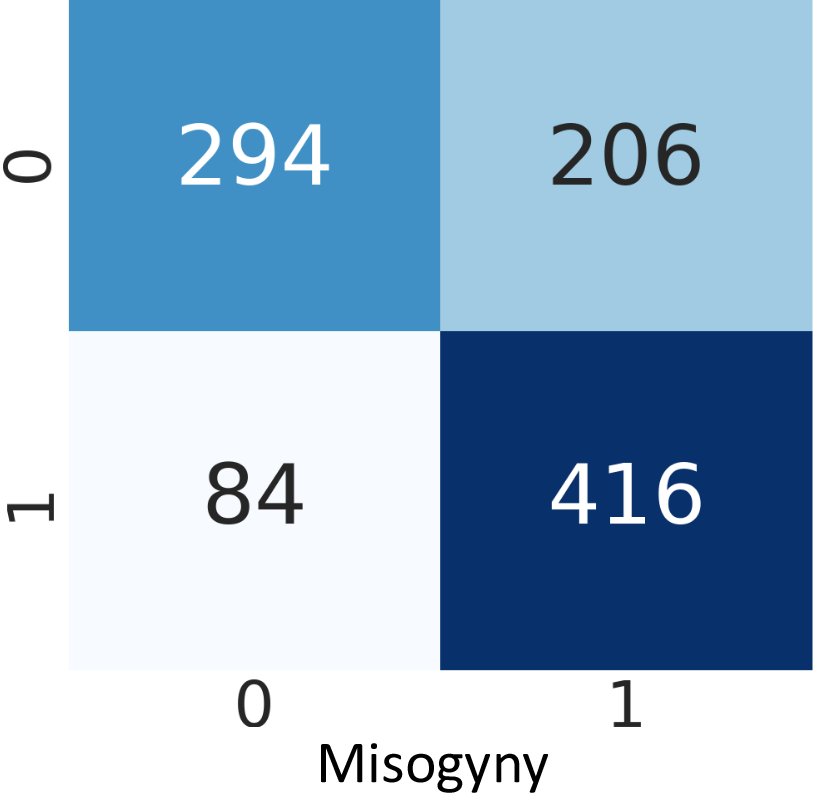}
    \includegraphics[width=3.1cm]{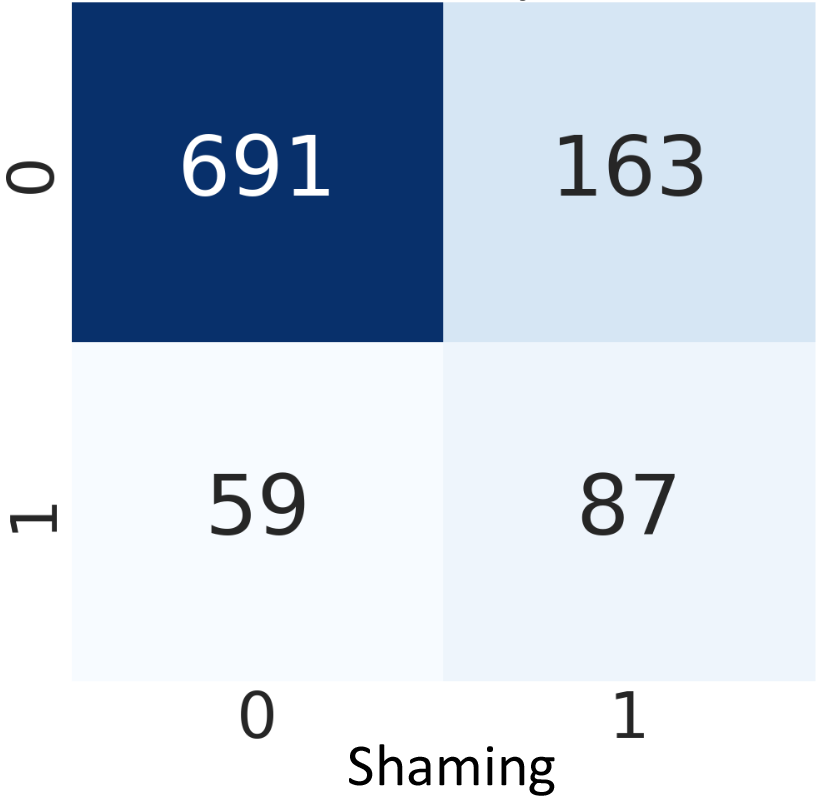}
    \includegraphics[width=3.1cm]{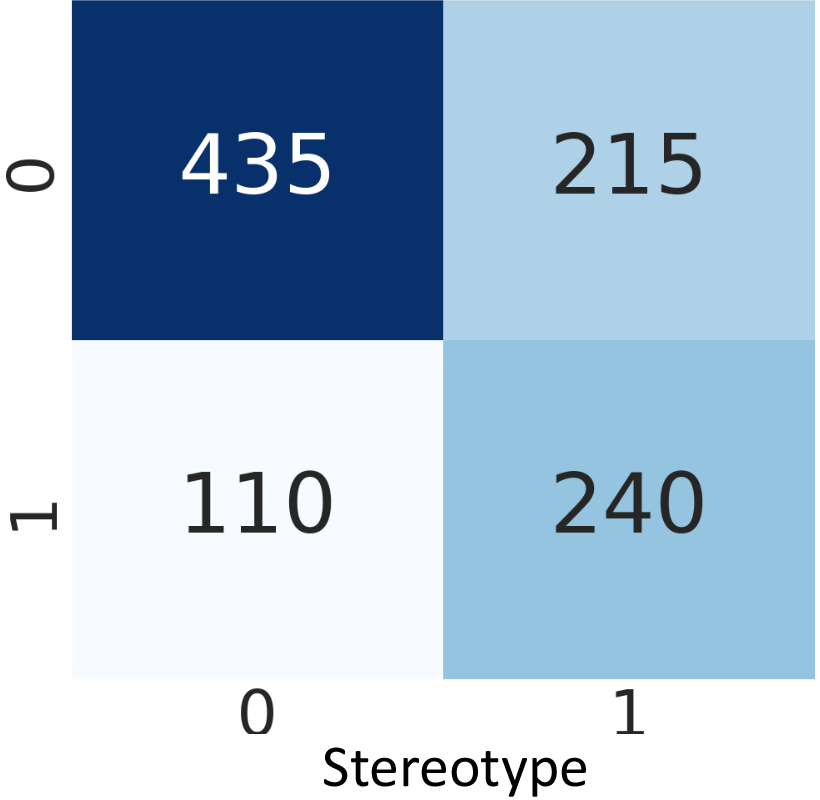}
    \includegraphics[width=3.1cm]{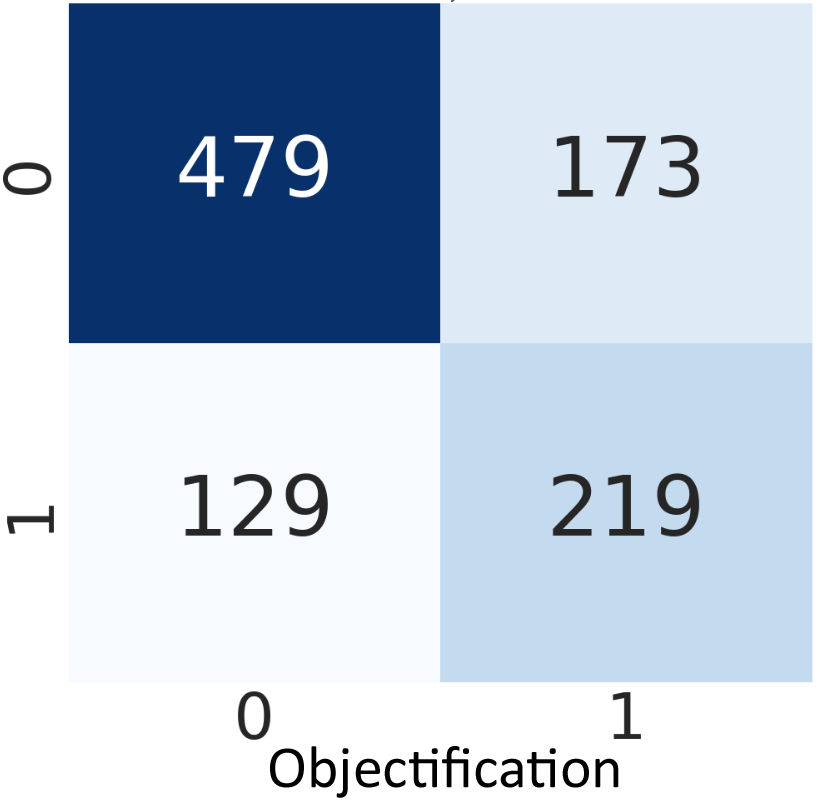}
    \includegraphics[width=3.1cm]{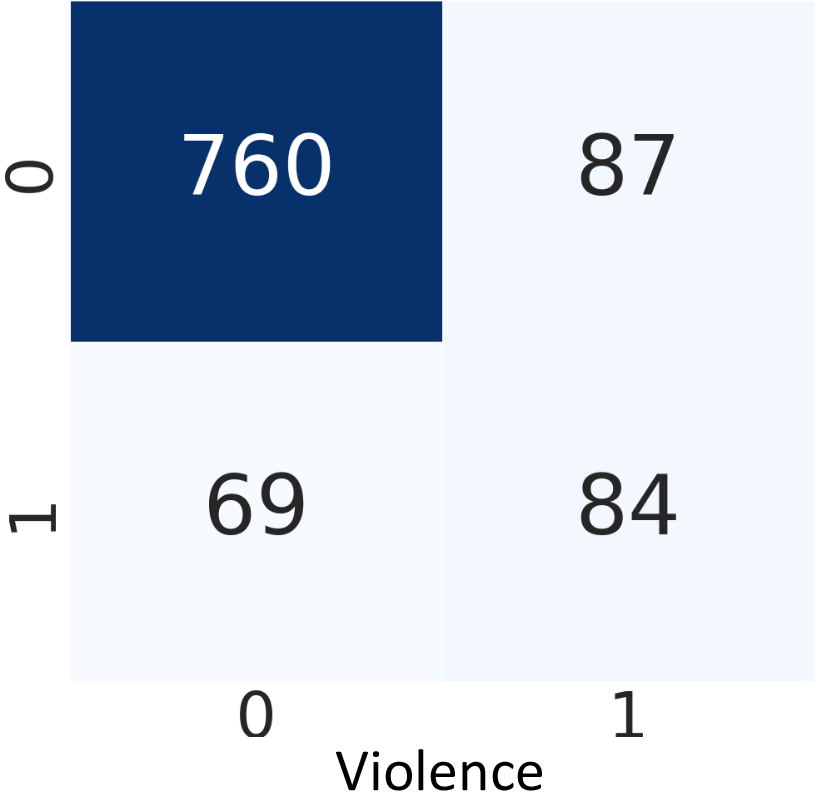}
    \caption{Per class confusion matrices for the ensemble model.}
    \label{fig:confusion-matrix}
\end{figure*}
 
Tables \ref{tab:resultsA} and \ref{tab:resultsB} are the best performances on the official competition test set for the models we trained. All modes were trained using the hyperparameters specified in the previous section.

Even though the performance on the binary classification from Sub-task A was comparable with the best single model (UNITER-VGCN), it was slightly lower. As seen in Figure \ref{fig:confusion-matrix}, our best system - the ensemble model - has the tendency to over-predict misogyny. Some erroneous predictions were driven by their aggressive language, for instance \textit{"IF YOU'RE DATING MY DAUGHTER AND YOUR STUPID ENOUGH TO DO THIS I'M GOING TO KILL YOU!" depicting also that a domestic abuse victim was understandably detected as misogynous type "violence"}. Records 15566 and 15311 depict a face close up of a famous woman and were  incorrectly labeled as misogynous, even if the text is replaced with a neutral or even positive one like "woman" or "best". This can be explained by the off-balance in the training data where the visual object "woman" is detected 2,262 times in the misogynous entries, and only 639 times in non-misogynous memes; similarly, "eyebrows" are four times more likely to appear in the misogynous class. However, memes like \textit{"When you know it's a trap but you can't wait to take the bait"} that depict a woman body are not detected as misogyny since these memes require additional background knowledge in order to understand the real intent.

\section{Conclusion}
In this paper, we describe the architectures used in our submission at Semeval-2022, Task 5: MAMI - Multimedia Automatic Misogyny Identification. Our proposed models took the pre-trained UNITER-base and UNITER-large models and enhanced them with image sentiment features or, by using a GCN, with additional domain information from an external dataset. Our best model achieved an F1-score of 71.4 for Sub-task A and 67.3 in Sub-task B, seen in Table \ref{tab:resultsTeams} in comparison to the top three results on each sub-task,  arguing that these models can perform reasonable well in a multi-modal setting and that the generalization power of the UNITER pre-trained model was enhanced by integrating image object nodes in the co-occurrence graph. Also, we showed that our ensemble smoothed out the uneven performance caused by different train/dev data splits and improved the overall performance. 

In terms of future work, we plan to continue our research into the automatic detection of abusive and hateful online content, and extend the experiments onto other pre-trained multi-modal models, as well as to attempt to improve their performance through task-adaptive pre-training \cite{gururangan2020don}. Even though Internet memes provide an interesting combination of textual message and image, they represent only one medium that can spread toxic messages. Studying other modalities like video or audio would help widen the understanding on how to detect and limit the spread of these undesired messages.

\bibliography{anthology,custom}

\end{document}